\documentclass[journal]{IEEEtran}
\usepackage{cite}
\usepackage{adjustbox}
\usepackage{amsmath}
\usepackage{graphicx}
\usepackage{textcomp}
\usepackage{csquotes}
\usepackage{adjustbox}
\usepackage{hyperref}
\usepackage{multirow}
\usepackage{titlesec}
\usepackage{subcaption}
\usepackage{makecell}
\usepackage{graphicx}
\usepackage{xcolor} 
\usepackage{amssymb}
\usepackage{makecell}
\usepackage{orcidlink}
\usepackage[ruled,vlined]{algorithm2e}
\usepackage{algpseudocode}

\usepackage{array}
\usepackage[numbers,sort&compress]{natbib}

\usepackage{etoolbox}
\usepackage{adjustbox}

\definecolor{columbiablue}{rgb}{0.61, 0.87, 1.0}
\titlespacing{\subsubsection}{0pt}{0pt}{1pt}

\usepackage{todonotes}
\usepackage{url}
\usepackage{gensymb}
\usepackage{bm}

\newcounter{algsubstate}

\newlength\myindent
\titleformat{\paragraph}
{\normalfont\normalsize\bfseries}{\theparagraph}{1em}{}
\titlespacing*{\paragraph}
{0pt}{3.25ex plus 1ex minus .2ex}{1.5ex plus .2ex}
\usepackage{times}
\usepackage[margin=0.8in]{geometry}
\usepackage{lipsum}
\usepackage{multicol}
\usepackage[table]{xcolor}
\usepackage{soul}
\usepackage{gensymb}
\usepackage{float}
\usepackage{authblk}
\usepackage{orcidlink}
\begin{document}
\title{A Hybrid Neural-Assisted Unscented Kalman Filter for Unmanned Ground Vehicle Navigation}
\author{Gal~Versano  \orcidlink{0009-0007-7766-2511},
        and~Itzik~Klein \orcidlink{0000-0001-7846-0654} ~\IEEEmembership{Senior Member,~IEEE}
        \thanks{G. Versano and I. Klein, are with the Autonomous Navigation and Sensor Fusion Lab, Hatter Department of Marine Technologies, Charney School of Marine Sciences, University of Haifa, Israel.}}
\maketitle
\begin{abstract}
\noindent Modern autonomous navigation for unmanned ground vehicles relies on different estimators to fuse inertial sensors and GNSS measurements. However, the constant noise covariance matrices often struggle to account for dynamic real-world conditions.
In this work we propose a hybrid estimation framework that bridges classical state estimation foundations with modern deep learning approaches. Instead of altering the fundamental unscented Kalman filter equations, a dedicated deep neural network is developed to predict the process and measurement noise uncertainty directly from raw inertial and GNSS measurements. We present a sim2real approach, with training performed only on simulative data. In this manner, we offer perfect ground truth data and relieves the burden of extensive data recordings. To evaluate our proposed approach and examine its generalization capabilities,  we employed a 160-minutes test set from three datasets each with different types of vehicles (off-road vehicle, passenger car, and mobile robot), inertial sensors, road surface, and environmental conditions. We demonstrate across the three datasets a position improvement of $12.7\%$ compared to the adaptive model-based approach. Thus, offering a scalable and a more robust solution for unmanned ground vehicles navigation tasks.
\end{abstract}
\section{Introduction}\label{intro_sec}
\noindent The rapid develop of autonomous ground vehicles is primarily driven by their potential to revolutionize operational efficiency, enhance industrial productivity, and provide flexible solutions for complex tasks
\cite{basiuk2023mobile, raj2022comprehensive,antonyshyn2023multiple}. Currently, autonomous systems are deployed across diverse sectors, including precision agriculture for automated harvesting, logistics for warehouse management, and high-risk environments such as search-and-rescue operations and underground exploration \cite{yepez2023mobile,lin2023intelligence,neaz2023design}. \\
%
\noindent To successfully complete their tasks, reliable, accurate, and continuous navigation is a fundamental mandatory requirement for unmanned ground vehicle (UGV) operations. To achieve high-frequency state estimation and improve system responsiveness, most autonomous platforms employ inertial sensors with external updates from different modalities. Common sensors include the global navigation satellite systems (GNSS), vision-based localization, and light detection and ranging (LiDAR) \cite{yue2024lidar,patoliya2022robust,li2022tightly}. \\
\noindent Due to the nonlinear nature of such fusion type, a nonlinear filter is required for the fusion process. 
The extended Kalman filter (EKF) has long been the most prevalent extension of the linear Kalman filter~\cite{fujii2013extended}. The EKF linearize the nonlinear state and measurement equations using a first-order Taylor expansions. However, local linearization often introduces significant errors in the estimated mean and covariance, potentially leading to filter instability or divergence in highly nonlinear scenarios \cite{brigadnov2023error}. To circumvent these issues, the unscented Kalman filter (UKF) was introduced \cite{wan2000unscented}. Instead of linearizing the functions, the UKF employs the unscented transform to propagate a set of deterministically chosen sigma points through the actual nonlinear dynamics. This approach captures the posterior mean and covariance with up to third-order Taylor series accuracy, offering superior performance in nonlinear scenarios such as complex vehicle dynamics and environmental conditions.  \\
\noindent The performance of any Kalman-based filter is heavily dependent on the accurate characterization of the process noise covariance matrix and the measurement noise covariance matrix \cite{gelb1974applied}, \cite{bar2011tracking}. However, in real-world conditions, these noise profiles are often time-varying and difficult to model analytically. Poorly tuned covariance matrices can lead to sub-optimal estimation or complete filter failure. Model-based adaptive filtering addresses this by dynamically adjusting the noise covariance online. A traditional approach is covariance matching, which utilizes innovation-based statistics to estimate noise levels in real time \cite{almagbile2010evaluating}. Building on this, the robust adaptive UKF modifies these matrices only when statistical inconsistencies are detected between the predicted and observed states \cite{zheng2018robust}. Other methods utilize sliding-window evaluations of residuals to refine navigation accuracy, particularly in specialized domains like underwater robotics \cite{iezzi2023adaptive}.\\
\noindent Recently, the integration of machine learning  and deep learning  has emerged as a powerful paradigm for inertial sensing and sensor fusion \cite{cohen2024inertial}, \cite{hurwitz2024deep}, \cite{etzion2025snake}, \cite{golroudbari2023recent}, \cite{he2023deep}. Several studies have proposed neural network architectures to regress the noise covariance parameters within nonlinear frameworks\cite{xiong2021q}. For instance, hybrid learning-based EKF models have been developed to tune the process noise based solely on raw inertial readings. Specific architectures, such as VIO-DualProNet, have been designed to estimate inertial noise within visual-inertial systems (VINS) by processing accelerometer and gyroscope data through dedicated networks \cite{solodar2024vio}. Furthermore, transformer-based models, such as the set-transformer, have been employed to learn time-varying process noise for Doppler velocity log (DLV)-aided inertial navigation \cite{cohen2025adaptive}.  Later, a closed-loop training approach that estimates the process and measurement noise covariances using the EKF was introduced in \cite{xu2024ekfnet}. Yet, such approaches require high computational effort in the training phase as the state dimensions increase. Using the UKF, Levy et al. \cite{levy2025adaptive} proposed an adaptive method in which the ProcessNet network is used to estimate the process noise covariance parameters for DVL/INS UKF fusion in autonomous underwater vehicle navigation using a real-to-real approach. Such data-driven approaches allow the filter to adapt to complex error characteristics that are often invisible to classical statistical methods.\\
\noindent In this work we propose an adaptive process and measurement neural network-assisted unscented Kalman filter, ANPMN-UKF, to adaptivity estimate the noise uncertainty for UGV navigation. Our approach offers a hybrid estimation framework that connects classical state estimation foundations with modern deep learning approaches. Instead of altering the fundamental unscented Kalman filter equations, a dedicated deep neural network is developed to predict the process and measurement noise uncertainty directly from raw inertial and GNSS measurements. In addition, we follow a sim2real approach, with training performed only on simulative data.\\
\noindent
Our main contributions are:
\begin{enumerate}
    \item \textbf{ANPMN-UKF Framework:}  
    We propose an adaptive process and measurement neural network-assisted unscented Kalman filter approach. It offers a learning-assisted UKF framework that adaptively estimates both the process noise covariance matrix and the measurement noise covariance matrix for UGV navigation.
    \item \textbf{Deep Learning Noise Estimator:}  
    A compact and computationally efficient deep neural network backbone architecture is designed to enable real-time operation and for dual usage. Building upon this backbone architecture, we offer $\sigma_{\textbf{Q}}$-Net to adaptively learn the inertial noise uncertainty for the process noise covariance and $\sigma_{\textbf{R}}$-Net to adaptively learn the GNSS position noise uncertainty for the measurement  noise covariance.
    \item \textbf{Sim2Real:}  
    Our neural networks is trained exclusively on simulated trajectories with known ground truth, allowing it to capture complex and nonlinear noise patterns in both the process and measurement models. In this manner, we offer perfect ground truth data and relieve the burden of extensive data recordings.
    \item \textbf{Multi Dataset Evaluations:}  
    The proposed approach is evaluated on three real-world datasets collected under different operating conditions, road surfaces, platforms (off-road vehicle, passenger car, and mobile robot), and inertial sensor types with a total of 160 minutes of recorded data. 
\end{enumerate}
\noindent Overall, our method achieves consistent positioning  accuracy gains across all evaluated datasets, with an average improvement of {22.70\%} over the  UKF and {12.72\%} over the model-based adaptive UKF. 
\noindent The rest of the paper is organized as follows: Section \ref{sec:pro} gives the inertial navigation and UKF equations. Section \ref{sec:ourpropsal}
presents our proposed approach including ANPMN-UKF and our sim2real method.  Section \ref{sec:results} presents the different datasets and the results of this study, Finally, Section \ref{sec:conclosuion} gives the conclusions.
\section{Problem Formulation} \label{sec:pro} 
\subsection{The UKF Algorithm} \label{ukf_sec}
\noindent Consider a nonlinear discrete-time system with an unobserved state vector
$\mathbf{x}_k$ governed by
\begin{equation}
\mathbf{x}_{k+1} = f(\mathbf{x}_k, \mathbf{w}_k)
\label{eq:state_model}
\end{equation}
where $\mathbf{x_k}$ is the state vector, $f(\cdot)$ is the nonlinear state transition function, and $\mathbf{w}_k$ denotes the process noise.

\noindent The measurement model is given by
\begin{equation}
\mathbf{z}_{k+1} = h(\mathbf{x}_{k+1}, \mathbf{v}_{k+1})
\label{eq:measurement_model}
\end{equation}
where $h(\cdot)$ is the nonlinear measurement function,
$\mathbf{v}_{k+1}$ is the measurement noise, and $\mathbf{z}_{k+1}$ is the
observed measurement at time step $k+1$.

\noindent The UKF is based on the scaled unscented transform (UT)~\cite{julier2002scaled}, which approximates the statistics of a random variable undergoing a nonlinear transformation by propagating a carefully chosen set of sigma points through the nonlinear function.

\noindent The UKF operates and consists of the following steps~\cite{wan2000unscented}: 
\subsubsection{Initialization}
\noindent 
Consider the initialized state vector $\mathbf{x}_0$ with a known mean and covariance:
\begin{align}
\hat{\mathbf{x}}_0 &= \mathbb{E}[{x}_0], \\
\mathbf{P}_0 &= \mathbb{E}\!\left[({x}_0 - \hat{\mathbf{x}}_0)
({x}_0 - \hat{\mathbf{x}}_0)^\mathsf{T}\right].
\end{align}
\noindent The weights for computing the mean (m upper case) and covariance (c upper case) are defined as:
\begin{align}
w_0^{(m)} &= \frac{\lambda}{n + \lambda}, \\
w_0^{(c)} &= \frac{\lambda}{n + \lambda} + (1 + \alpha^2 + \beta), \\
w_i^{(m)} &= w_i^{(c)} = \frac{1}{2(n + \lambda)}, \quad i = 1, \ldots, 2n,
\end{align}
where $n$ is the dimension of the state vector and
$\lambda = \alpha^2 (n + \kappa) - n$.
The parameters $\alpha$, $\beta$, and $\kappa$ control the spread and weighting of the sigma points and are addressed as hyperparameters. Typically, $\alpha = 10^{-3}$, $\kappa = 0$, and $\beta = 2$ for
Gaussian distributions~\cite{nielsen2021ukf}.
\subsubsection{Sigma Points}
\noindent The sigma points are constructed as
\begin{align}
\boldsymbol{\chi}_{0,k} &= \hat{\mathbf{x}}_{k}, \quad i=0 \\
\boldsymbol{\chi}_{i,k} &= \hat{\mathbf{x}}_{k}
+ \left[\sqrt{(n+\lambda)\mathbf{P}_{k}}\right]_i,
\quad i = 1, \ldots, n, \\
\boldsymbol{\chi}_{i,k} &= \hat{\mathbf{x}}_{k}
- \left[\sqrt{(n+\lambda)\mathbf{P}_{k}}\right]_{i-n},
\quad i = n+1, \ldots, 2n,
\end{align}
\noindent where $\sqrt{\mathbf{P}_{k}}$ denotes the Cholesky decomposition of the covariance matrix.
\subsubsection{Prediction step}
\noindent Each sigma point is propagated through the nonlinear model:
\begin{equation}
\boldsymbol{\chi}_{i,k+1|k} = f(\boldsymbol{\chi}_{i,k}), \quad i = 0, \ldots, 2n.
\end{equation}

\noindent The predicted state mean and covariance are then computed as
\begin{equation}
\hat{\mathbf{x}}_{k+1|k}^{-} =
\sum_{i=0}^{2n} w_i^{(m)} \boldsymbol{\chi}_{i,k+1|k}    
\end{equation}
\begin{equation}
\begin{aligned}
\mathbf{P}_{k+1|k}^{-} &=
\sum_{i=0}^{2n} w_i^{(c)}
(\boldsymbol{\chi}_{i,k+1|k} - \hat{\mathbf{x}}_{k+1|k})
(\boldsymbol{\chi}_{i,k+1|k} - \hat{\mathbf{x}}_{k+1|k})^\mathsf{T} \\
&\quad + \mathbf{Q}_{k+1}.
\end{aligned}
\end{equation}

\noindent where $\mathbf{Q}_{k+1}$ is the covariance process noise matrix in time stamp k+1.
\subsubsection{Update step}
\noindent A new set of sigma points is generated from the predicted mean and covariance and propagated through the measurement function:
\begin{align}
\mathbf{\mathcal{Z}}_{i,k+1|k} &= h(\boldsymbol{\chi}_{i,k+1|k}), \quad i = 0, \ldots, 2n.
\end{align}

\noindent The predicted measurement mean and covariance are computed as:

\begin{equation}
\hat{\mathbf{z}}_{k+1|k} =
\sum_{i=0}^{2n} w_i^{(m)} \mathbf{\mathcal{Z}}_{i,k+1|k}
\end{equation}

\begin{align}
\mathbf{S}_{k+1}
&=
\sum_{i=0}^{2n} w_i^{(c)}
\bigl(
\mathbf{\mathcal{Z}}_{i,k+1|k} - \hat{\mathbf{z}}_{k+1|k}
\bigr)
\bigl(
\mathbf{\mathcal{Z}}_{i,k+1|k} - \hat{\mathbf{z}}_{k+1|k}
\bigr)^{\mathsf{T}}
\nonumber\\
&\quad + \mathbf{R}_{k+1}.
\label{innovation}
\end{align}

\noindent where $\mathbf{R}_{k+1}$ is the covariance measurement matrix noise in time stamp k+1.

\noindent The cross-covariance matrix is define by:
\begin{equation}
\mathbf{P}_{xz,k+1} =
\sum_{i=0}^{2n} w_i^{(c)}
(\boldsymbol{\chi}_{i,k+1|k} - \hat{\mathbf{x}}_{k+1|k})
(\mathbf{\mathcal{Z}}_{i,k+1|k} - \hat{\mathbf{z}}_{k+1|k})^\mathsf{T}
\label{cross_covariance}
\end{equation}

\noindent The Kalman gain is computed by using \eqref{innovation} and \eqref{cross_covariance}:
\begin{equation}
\mathbf{K}_{k+1} = \mathbf{P}_{xz,k+1} \mathbf{S}_{k+1}^{-1}
\label{kalman_gain}
\end{equation}
\noindent Finally, the state mean and covariance are updated using the measurement $\mathbf{z}_{k+1}$ and the Kalman gain \eqref{kalman_gain}:
\begin{align}
\hat{\mathbf{x}}_{k+1}^{+} &=
\hat{\mathbf{x}}_{k+1|k}^{-} +
\mathbf{K}_{k+1}(\mathbf{z}_{k+1} - \hat{\mathbf{z}}_{k+1|k}) \\
\mathbf{P}_{k+1}^{+} &=
\mathbf{P}_{k+1|k}^{-} -
\mathbf{K}_{k+1} \mathbf{S}_{k+1} \mathbf{K}_{k+1}^\mathsf{T}
\end{align}
\subsection{INS/GNSS Fusion} \label{ins_gnss_fus}
\noindent The INS/GNSS fusion is considered in the
error-state implementation.
The system state is decomposed into a nominal state propagated by the INS and an error state estimated by the UKF. The 15-dimensional error-state vector is defined as:
\begin{equation}
\delta \mathbf{x} =
\begin{bmatrix}
\delta \mathbf{p}^n  \hspace{1.5 mm}
\delta \mathbf{v}^n \hspace{1.5 mm}
\delta \boldsymbol{\psi}^n \hspace{1.5 mm}
\mathbf{b}_a^b  \hspace{1.5 mm}
\mathbf{b}_g^b \hspace{1.5 mm}
\end{bmatrix}{^T}
\in \mathbb{R}^{15 \times 1}
\end{equation}
where $\delta \mathbf{p}^n\in\mathbb{R}^3$ is the position vector error state expressed in the navigation frame,  $\delta\mathbf{v}^n\in \mathbb{R}^3$, is the velocity vector error state expressed in the navigation frame, $\delta\boldsymbol{\psi}^n\in\mathbb{R}^3$ is the misalignment error expressed in the navigation frame, $\mathbf{b}_a\in \mathbb{R}^3$ is the residual bias of the accelerometers expressed in the body frame, and $\mathbf{b}_g\in\mathbb{R}^3$ is the residual bias of the gyroscopes expressed in the body frame.
\subsubsection{Error-State Dynamics}
\noindent
The UKF continuous-time error-state dynamics are described by \cite{farrell2008aided}:
\begin{equation}
\dot{\delta \mathbf{x}} =
\mathbf{F}\,\delta \mathbf{x} +
\mathbf{G}\,\delta \mathbf{w}
\end{equation}
where $\mathbf{F}$ is the system matrix derived from the nonlinear INS kinematics, and $\mathbf{G}$ is the noise distribution matrix. The process noise vector includes the inertial sensors measurement noise and the random-walk behavior of the sensor biases:
\begin{equation}
\delta  \mathbf{w} =
\begin{bmatrix}
\mathbf{n}_a^\mathsf{T} &
\mathbf{n}_g^\mathsf{T} &
\mathbf{n}_{ab}^\mathsf{T} &
\mathbf{n}_{gb}^\mathsf{T}
\end{bmatrix}^\mathsf{T} \in \mathbb{R}^{12\times 1}
\end{equation}
where $\mathbf{n}_a \in \mathbb{R}^3$ and $\mathbf{n}_g \in \mathbb{R}^3$ are the measurement noise of the accelerometers and gyroscopes, respectively, and $\mathbf{n}_{ab} \in \mathbb{R}^3$ and $\mathbf{n}_{gb} \in \mathbb{R}^3$ are the noise terms of the accelerometer and gyroscope bias random-walk models, respectively.
\noindent The discrete process noise covariance matrix is obtained based on the commonly used approximation \cite{titterton2004strapdown}:
\begin{equation}
    \mathbf{Q}_{{k+1}}^{d}=\mathbf{G}_{k+1}\mathbf{Q}_{c_{k+1}}\mathbf{G}_{k+1}^{T}\Delta t \in \mathbb{R}^{15\times15}
    \label{dis_Q}
\end{equation}
\noindent where the continuous process noise covariance matrix is:
\begin{equation}
    \mathbf Q_{c_{k+1}}=diag(\mathbf{\delta w}) \in \mathbb{R}^{12 \times 12}
\end{equation}
\subsubsection{Measurement Model}
\noindent  GNSS position updates in the geodetic frame $(\phi, \lambda, h)$ are used to update the filter. In an  UKF framework, the predicted measurement $\hat{\mathbf{z}}_{k+1|k}$ is computed by propagating sigma points $\mathbf{x}_{i,k+1|k}$ through the observation function $h(\cdot)$:
\begin{equation}
\hat{\mathbf{z}}_{k+1|k} = \sum_{i=0}^{2n} w_i^{(m)} h(\mathbf{x}_{i, k+1|k})
\end{equation}
\subsection{Model-Based Adaptive Noise Covariance Estimation} \label{model_base_adp}
\noindent In UKF formulation, the process noise covariance matrix
$\mathbf{Q}_{d_{k}}$ is assumed to be constant and is tuned offline for an optimal value. While this assumption simplifies filter design, it limits performance in practical navigation scenarios where sensor noise characteristics and environmental conditions vary over time and thus result in a varying process noise covariance. \\
\noindent 

\begin{figure*}
    \centering
    \includegraphics[width=1\linewidth]{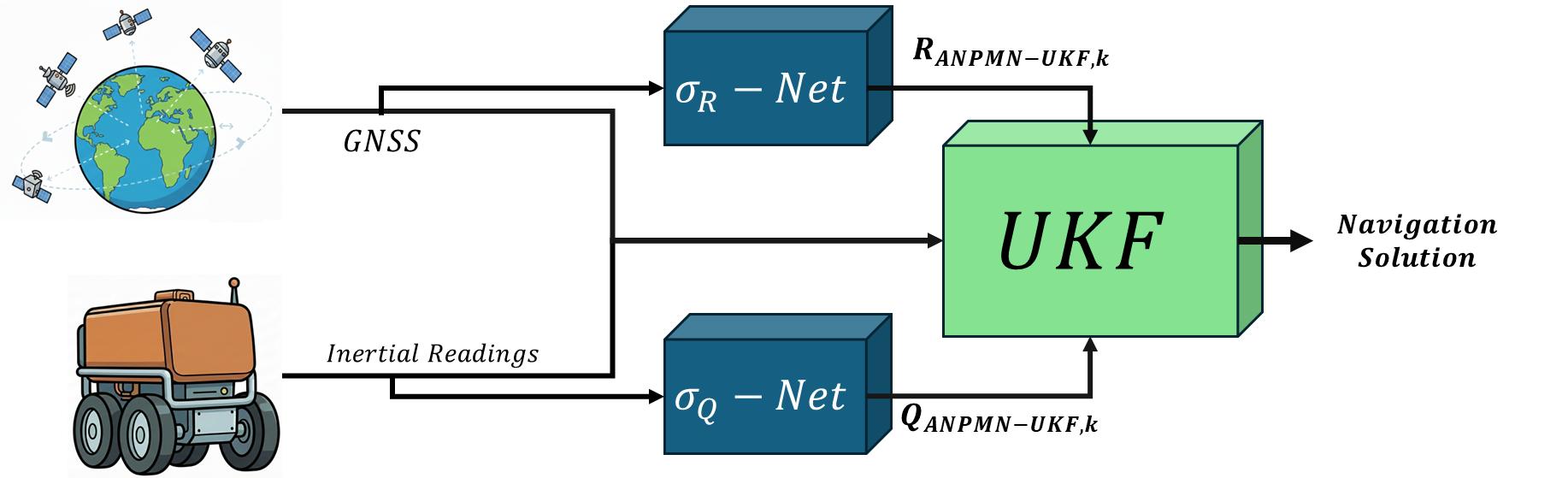}
    \caption{Our proposed ANPMN-UKF implemented on the GNSS/INS fusion scansion for unmanned ground vehicles.}
    \label{fig:approach}
\end{figure*}
\noindent
One of the most common approaches to update the discrete process noise covariance matrix $\mathbf{Q}_{k}$ during operation is model-based adaptive strategy based on \cite{mohamed1999adaptive} is adopted to update the discrete process noise covariance matrix $\mathbf{Q}_{k}$ during operation. 
This approach is based on the innovation sequence by calculating the innovation covariance matrix over a sliding window of length $\eta$ as
\begin{equation}
\mathbf{C}_{k} =\frac{1}{\eta} \sum_{i=k-\eta+1}^{k}
\boldsymbol{\nu}_{i}\boldsymbol{\nu}_{i}^{\mathsf{T}},
\label{eq:innovation_cov}
\end{equation}
where $\boldsymbol{\nu}_{i}$ denotes the innovation vector at time step $i$ defined by:
\begin{equation}
\boldsymbol{\nu}_{k+1} = \mathbf{z}_{k+1} - \sum_{i=0}^{2n} w_i^{(m)} \mathbf{\mathcal{Z}}_{i,k+1|k}
\label{eq:innovation}
\end{equation}
with $\mathbf{z}_{k}$ being the measurement vector,
and $\sum_{i=0}^{2n} w_i^{(m)} \mathbf{\mathcal{Z}}_{i,k+1|k}$ is the predicted state estimation.
The estimated innovation covariance $\mathbf{C}_{k}$ is then utilized to update the process noise covariance according to
\begin{equation}
\hat{\mathbf{Q}}_{k}
=
\mathbf{K}_{k}\,
\mathbf{C}_{k}\,
\mathbf{K}_{k}^{\mathsf{T}}
\label{eq:Q_adapt}
\end{equation}
where $\mathbf{K}_{k}$ is the Kalman gain at time step $k$.

\noindent The measurement noise covariance can also be made adaptive to cope with the external sensor varying conditions. In the UKF, the predicted measurement covariance is computed as
\begin{equation}
\mathbf{S}^-_k =
\sum_{i=0}^{2n} w_i^{(c)}\left(\mathbf{\mathcal{Z}}_{i,k|k-1} - \hat{\mathbf{z}}_{k|k-1}
\right)\left(\mathcal{Z}_{i,k|k-1} - \hat{\mathbf{z}}_{k|k-1}
\right)^\mathsf{T}
\end{equation}
\noindent where $\mathcal{Z}_{i,k|k-1}$ denotes the predicted measurement corresponding to the $i$-th sigma point.

\noindent Using these quantities, an innovation-based adaptive estimate of the measurement noise covariance is obtained as:
\begin{equation}
\hat{\mathbf{R}}_k ={\mathbf{C}}_k - \mathbf{S}^-_k
\end{equation}
\section{Proposed Approach} \label{sec:ourpropsal} 
\noindent 
To cope with the varying process and measurement noise covariance values during real world conditions, we offer ANPMN-UKF an adaptive neural process and measurement noise approach to estimate the those covariances and plug them into the well established UKF. We offer the same neural architecture backbone for estimating the inertial uncertainty, $\sigma_{\textbf{Q}}$-Net, and for estimating the position uncertainty, $\sigma_{\textbf{R}}$-Net. \\
We follow a sim2real approach for training on simulative data while testing on real recoded data. To this end, simulated trajectories are generated to train the neural networks on different standard deviations of inertial sensor noise emulating the varying process noise. Specifically, we consider the noise in the specific force and angular velocity measurements, $[\mathbf{n}_a, \mathbf{n}_g] \in \mathbb{R}^{6\times 1}$. In the same manner, to cope the varying measurement noise, we focus on GNSS position updates and train a different neural network on different values of the position standard deviation $[\mathbf{n}_p] \in \mathbb{R}^{3 \times 1}$. The trained networks are then used in real world implementation to adaptively estimate the process and measurement noise covariances used in the UKF framework. Our proposed approach is illustrated in Figure~\ref{fig:approach}. In the next sections, we elaborate on each component of our proposed approach.
\subsection{Simulated Data Generation} \label{sim_data}
\begin{figure}[h]
    \centering
    \includegraphics[width=0.95\linewidth]{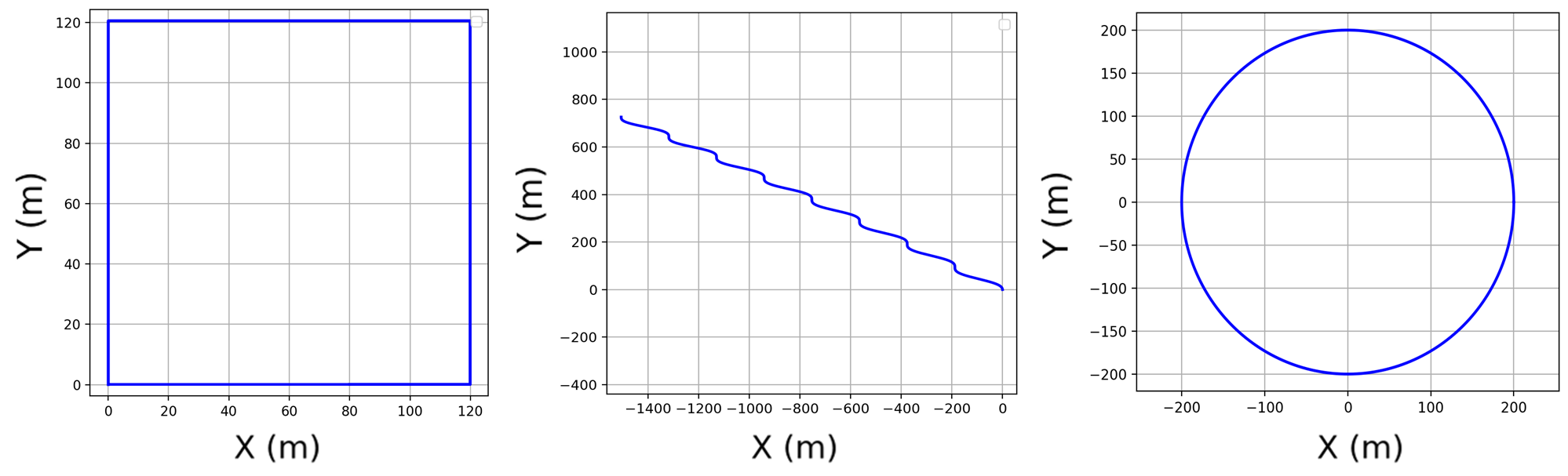}
    \caption{Examples of different simulated trajectories in 2D-view showing rectangular, sine wave, and circle trajectories.}
    \label{fig:sim_2D_traj}
\end{figure}

\noindent  Motivated by the success of using simulated data for process noise covariance estimation in quadrotor and autonomous underwater vehicles  dynamics~\cite{or2022hybrid, or2022hybrid2, or2023pronet}, we adopt a similar approach for unmanned ground vehicles dynamics and elaborate it also for the regression of the measurement noise covariance. 
The core idea is that the diversity of simulated trajectories allows the neural model to generalize effectively to unseen real-world scenarios.
The simulated dataset generation was created by constructing five different baseline trajectories, each with different dynamics including a straight line, rectangular, circle, and a sine wave. An example of 2D-view of the trajectories is shown in Figure~\ref{fig:sim_2D_traj}. The diversity of these trajectories introduces sufficient motion variability, enabling the learned model to generalize to previously unseen trajectories.
\\
\noindent The baseline trajectories' duration was between 100 and 500 seconds. There, we generated ideal inertial readings (specific force and angular velocity vectors) and ground truth (GT) position vectors at a sampling rate of 100[Hz]. This gives $T\times 100$ samples for each trajectory with $T$ standing for the duration of the trajectory. Next, zero mean white Gaussian noise was added to the inertial readings and position measurements. For the inertial sensor readings in each of the six axes (three accelerometers and three gyroscopes) zero mean white Gaussian noise with standard deviation values in the range $[\boldsymbol{\sigma}_a, \boldsymbol{\sigma}_g] \in [0.001,0.02]$ was added in each sample. Starting with the lower value until the maximum value. This procedure was repeated 25 times (equal step size) resulting in  $T\times 100 \times 25$ samples for each inertial axis. Thus, the measured specific force and angular velocity vectors in sample $i$ of trajectory $j$ with noise level $k$ are
\begin{eqnarray}
 \tilde{\mathbf{f}}^{b}_{(j,i)} & = & \mathbf{f}^{b}_{\text{GT},(j,i)}+\mathbf{n}_{a, (j,i)},  \hspace{1.5 mm} \mathbf{n}_a \sim \mathcal{N}(0,\boldsymbol{\sigma}_{a,(j,k)}^2) \\
 \tilde{\boldsymbol{\omega}}^{b}_{(j,i)} & = & \boldsymbol{\omega}^{b}_{\text{GT},(j,i)}+\mathbf{n}_{g, (j,i)},  \hspace{1.5 mm} \mathbf{n}_g \sim \mathcal{N}(0,\boldsymbol{\sigma}_{g,(j,k)}^2)
\end{eqnarray}
In the same manner, zero mean white Gaussian noise with standard deviation values in the range $[\boldsymbol{\sigma}_p] \in [1.5,3.0]$[m] was added in each sample of the position measurements creating $T\times 100 \times 25$ measurements for each trajectory. Thus, the measured position in sample $i$ of trajectory $j$ with noise level $k$ is
\begin{equation}
 \tilde{\mathbf{p}}^{n}_{(j,i)}=\mathbf{p}^{n}_{\text{GT},(j,i)}+\mathbf{n}_{p, (j,i)},  \hspace{1.5 mm} \mathbf{n}_p \sim \mathcal{N}(0,\boldsymbol{\sigma}_{p,(j,k)}^2)
\end{equation}
The entire process of generating the simulated data is illustrated in Algorithm~\ref{alg:sim_dataset_generation_jik}. 
\begin{algorithm}[t]
\caption{Simulated Dataset Generation}
\label{alg:sim_dataset_generation_jik}
\DontPrintSemicolon

\SetKwInOut{Input}{Input}
\SetKwInOut{Output}{Output}

\Input{
    Baseline trajectories $\mathcal{T}=\{\tau_j\}_{j=1}^{5}$\\
    Sampling rate $f_s=100$ [Hz]\\
    Trajectory duration range $T \in [100,500]$ [sec]\\
    Number of noise levels $K=25$\\
    Inertial noise ranges $\sigma_a, \sigma_g \in [0.001,0.02]$\\
    Position noise range $\sigma_p \in [1.5,3.0]$\\
}
\Output{
    Dataset $\mathcal{D}$ of noisy measurements and noise labels\;
}

\textbf{Initialize:} $\mathcal{D} \leftarrow \emptyset$\\
Construct uniform grids for noise levels: $\{\sigma_{a,k}, \sigma_{g,k}, \sigma_{p,k}\}_{k=1}^{K}$\;

\For{$j=1$ \KwTo $J$}{
    Generate ideal sequences at $f_s$: 
    $\{\mathbf{f}^{b}_{\text{GT},i}, \boldsymbol{\omega}^{b}_{\text{GT},i}, \mathbf{p}^{n}_{\text{GT},i}\}_{i=1}^{N_j}$\;

    \For{$k=1$ \KwTo $K$}{
        \For{$i=1$ \KwTo $N_j$}{
             $\tilde{\mathbf{f}}^{b}_{(j,i)} \leftarrow \mathbf{f}^{b}_{\mathrm{GT},(j,i)} + \mathbf{n}_{a,(j,i)}$\\
            $\tilde{\boldsymbol{\omega}}^{b}_{(j,i)} \leftarrow \boldsymbol{\omega}^{b}_{\mathrm{GT},(j,i)} + \mathbf{n}_{g,(j,i)}$\\
            $\tilde{\mathbf{p}}^{n}_{(j,i)} \leftarrow \mathbf{p}^{n}_{\mathrm{GT},(j,i)} + \mathbf{n}_{p,(j,i)}$\\
            $\mathcal{D} \leftarrow 
            \left\{\tilde{\mathbf{f}}^{b}_{(j,i)},\tilde{\boldsymbol{\omega}}^{b}_{(j,i)},\tilde{\mathbf{p}}^{n}_{(j,i)};
            \boldsymbol{\sigma}_{a,(j,k)},\boldsymbol{\sigma}_{g,(j,k)},\boldsymbol{\sigma}_{p,(j,k)}\right\}$\;
        }

    }
}
\Return $\mathcal{D}$\;
\end{algorithm}
%
\begin{figure*}[h]
    \centering
    \includegraphics[width=1\linewidth]{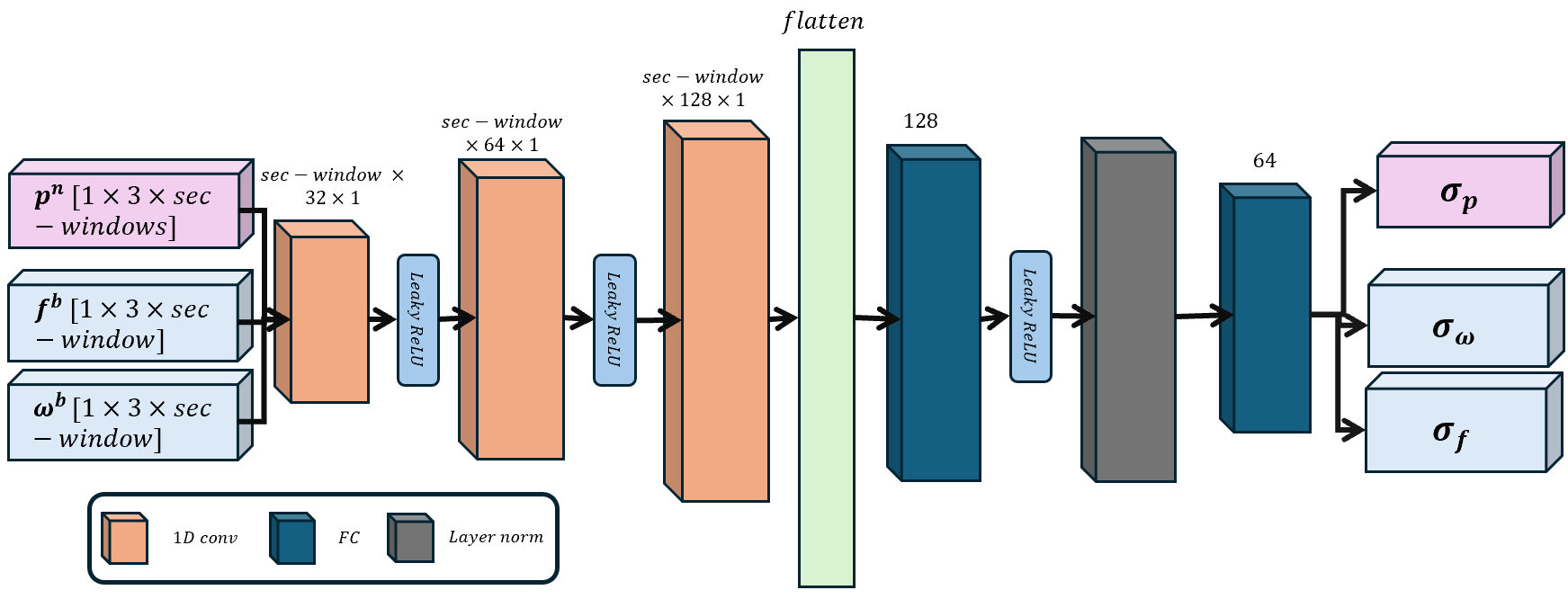}
    \caption{Our dual-usage deep neural network architecture for the estimation of the inertial sensor measurement  (input/output in light blue) and position measurement  (input/output in pink) standard deviations.}
    \label{DNN_archi}
\end{figure*}
\subsection{DNN Architecture}
\noindent Our deep dual-usage deep neural network (DNN) backbone architecture is used for the estimation of the inertial sensor measurement uncertainty, $\sigma_{\textbf{Q}}$-Net and the position measurement uncertainty $\sigma_{\textbf{R}}$-Net. For each task the same backbone is used with different input and output.
The backbone architecture is composed of one-dimensional convolutional layers, followed by fully connected layers and layer normalization, with the full architecture shown in Fig,\ref{DNN_archi}. Our network design leverages convolutional layers' robust temporal feature extraction capabilities, avoiding excessive sensitivity to sequential orders observed in other time series architectures like recurrent neural networks. This property is useful in addressing the problem of noise covariance, in which dependencies and patterns across data are important factors. Moreover, this architecture offers a simple yet efficient solution to the regression problem at hand.
\subsection{Training Procedure}
\noindent For training, the GT standard deviations are first computed, as described in Section~\ref{sim_data}. The networks are trained using sliding windows consisting of sensor data. For $\sigma_{\textbf{Q}}$-Net, the input consists of six channels (three accelerometer and three gyroscope axes) with a window length of one second consisting of 100 measurements in each channel. 
For $\sigma_{\textbf{R}}$-Net, the input consists of three channels (for each position component) with a window length of one second consisting of 100 measurements in each channel. 
For both inertial and position networks, a batch size of 64 is used for training.\\
\noindent The output of the network is a vector of estimated standard deviations, with dimensionality of six for $\sigma_{\textbf{Q}}$-Net and three for the $\sigma_{\textbf{R}}$-Net. The networks are trained using the mean squared error (MSE) loss function, defined as
\begin{equation}
\mathrm{MSE}(\boldsymbol{\sigma}_{\text{GT}}, \hat{\boldsymbol{\sigma}})
=
\frac{1}{N}
\sum_{i=1}^{N}
\left( \sigma_{\text{GT},i} - \hat{\sigma}_i \right)^2,
\end{equation}
where $\boldsymbol{\sigma}_{\text{GT}}$ denotes the GT standard deviation vector, $\hat{\boldsymbol{\sigma}}$ is the network output, and $N$ is the number of samples. 

\noindent During inference, the input data are processed through the network layers via forward propagation. The network parameters are learned using a stochastic gradient descent–based optimization scheme, where the weights and biases are updated according to
\begin{equation}
\boldsymbol{\theta} = \boldsymbol{\theta} - \eta \nabla_{\boldsymbol{\theta}} J(\boldsymbol{\theta}),
\end{equation}
where $\boldsymbol{\theta} = [\boldsymbol{\omega}, \mathbf{b}]^{\mathsf{T}}$ denotes the vector of all network weights and biases, $J(\boldsymbol{\theta})$ is the loss function, $\eta$ is the learning rate, and $\nabla_{\boldsymbol{\theta}}$ is the gradient operator.
In the proposed method, the DNN model is trained using the adaptive moment estimation (Adam) optimizer to improve convergence speed and stability. The main hyperparameters used during training are summarized in Table~\ref{tab:hyperp}.
\begin{table}[h] 
\centering
\caption{Main training hyperparameters values.}
\begin{tabular}{|c|c|c|c|}\hline
Learning Rate & Batch Size & Epochs & Window Size \\
\hline
0.001 & 64 & 200 & Sec window IMU \\
\hline
\end{tabular} \label{tab:hyperp}
\end{table}
\subsection{Process and Measurement Noise Covariance} \label{dnn_test}
\noindent The trained networks outputs the inertial uncertainty and the position  uncertainty used to create the adaptive covariance matrix.  The inertial standard deviation from $\sigma_{\textbf{Q}}$-Net is used to create the process noise covariance matrix $\mathbf{Q}_{DNN}$ while the position standard deviation from $\sigma_{\textbf{R}}$-Net is used to create the measurement noise covariance matrix $\mathbf{R}_{DNN}$ at each time step. These estimated covariances are then incorporated into the UKF framework is described in Section~\ref{sec:ourpropsal}.

\noindent The continuous-time process noise covariance matrix, based on $\sigma_{\textbf{Q}}$-Net, is defined as
\begin{equation}
\mathbf{Q}_{\text{ANPMN},k} = \mathrm{diag}
\begin{bmatrix}
\mathbf{n}_{a_{\textbf{Q}-Net}}^\mathsf{T} &
\mathbf{n}_{g_{\textbf{Q}-Net}}^\mathsf{T} &
\mathbf{n}_{ab}^\mathsf{T} &
\mathbf{n}_{gb}^\mathsf{T}
\end{bmatrix}^\mathsf{T}
\in \mathbb{R}^{12 \times 12}
\label{eq:Q_dnn}
\end{equation}
where $\mathbf{n}_{a_{\textbf{Q}-Net}}$, $\mathbf{n}_{g_{\textbf{Q}-Net}}$, correspond to the accelerometer noise, gyroscope noise estimated by $\sigma_{\textbf{Q}}$-Net, respectively.
The remaining terms, {$\mathbf{n}_{ab}$} and {$\mathbf{n}_{gb}$}, represent the accelerometer and gyroscope bias random-walk noises. 
In this work, they are kept fixed and are not estimated by the network, since these bias processes change slowly and are difficult to identify reliably from the available measurements. Therefore, the network adapts only the fast time-varying inertial noise terms, which have the largest impact on filter performance.\\
\noindent The measurement noise covariance matrix at time $k$ is constructed using  $\sigma_{\textbf{R}}$-Net:
\begin{equation}
\mathbf{R}_{\text{ANPMN},k} = \mathrm{diag}
\begin{bmatrix}
\sigma^{2}_{N} &
\sigma^{2}_{E} &
\sigma^{2}_{D}
\end{bmatrix}
\in \mathbb{R}^{3 \times 3}
\label{eq:R_dnn}
\end{equation}
where $\sigma_{N}$, $\sigma_{E}$, and $\sigma_{D}$ are the regressed noise standard deviation in the north, east, and down directions, respectively.

\noindent The continuous-time process noise covariance in~\eqref{eq:Q_dnn} is discretized at each UKF prediction step using the method described in~\eqref{dis_Q} to obtain the discrete-time covariance matrix $\mathbf{Q}_{d,k}$.

\noindent Finally, weighting parameters $\alpha$ and $\beta$ are introduced to blend the ANPMN-estimated covariance matrices with predefined constant covariance matrices, yielding the final covariances used in our proposed ANPMN-UKF:
\begin{equation}
\mathbf{Q}_{\mathrm{UKF},k}
=
\alpha_{\text{ANPMN}}\,\mathbf{Q}_{\text{ANPMN},k}^{d}
+
(1-\alpha_{\text{ANPMN}})\,\mathbf{Q}_{k}^{d}
\end{equation}

\begin{equation}
\mathbf{R}_{UKF,k}
=
\beta_{\text{ANPMN}}\,\mathbf{R}_{\text{ANPMN},k}
+
(1-\beta_{\text{ANPMN}})\,\mathbf{R}_{\mathrm{}k}
\end{equation}
The parameters $\alpha_{\text{ANPMN}}$ and $\beta_{\text{ANPMN}}$ were determined using a grid search to optimize the blending between adaptive neural estimates and covariance matrices. The values are presented in Table \ref{values_dnn} 
\begin{table}[h]
\caption{Values of tuning parameters.}
\centering
\begin{tabular}{|c|c|}
\hline
 \textbf{Parameters}                              & \textbf{Value} \\ \hline
\textbf{$\alpha_{\text{ANPMN}}$} & 0.5            \\ \hline
\textbf{$\beta_{\text{ANPMN}}$}  & 0.7            \\ \hline
\end{tabular}
\label{values_dnn}
\end{table}

\section{Analysis and results} \label{sec:results} 
\subsection{Test Datasets}
\begin{figure*}[h]
    \centering
    \includegraphics[width=1\linewidth]{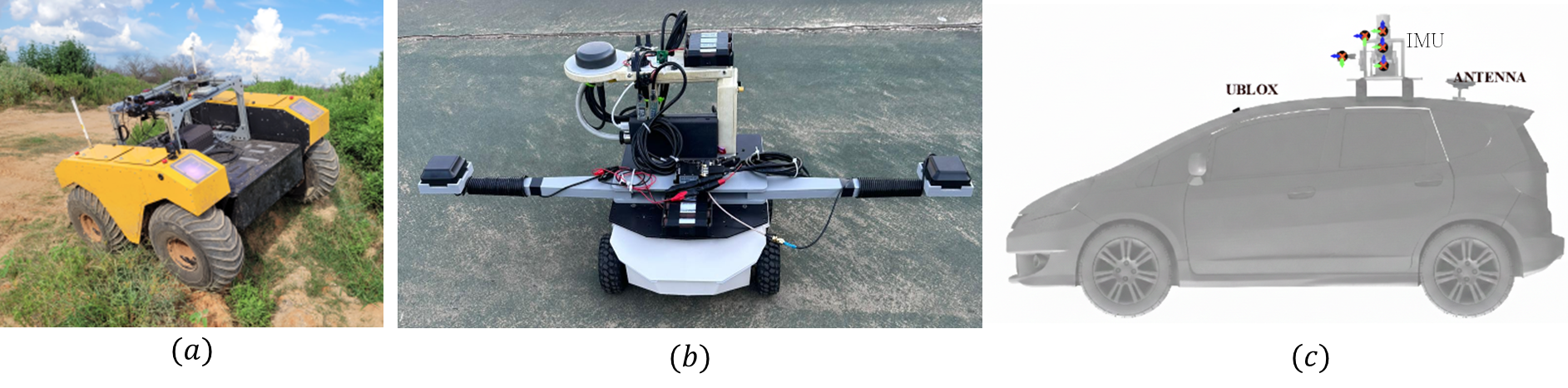}
    \caption{(a) ROOAD dataset \cite{chustz2021rooad} with the Waterhod-UGV platform\cite{clearpath_warthog}, (b) Our recorded dataset using the ROSbot platform~\cite{husarion_rosbot_xl_manual} and Arazim's IMU EX-300 \cite{arazim_ex300} , (c) Hong-Kong dataset \cite{hsu2023hongkong} recorded using a car.}
    \label{platform_datasets}
\end{figure*}

\noindent To evaluate the proposed algorithm, we employed three real-world datasets from different mobile platforms: two are publicly available, while the third was recorded specifically for this research. 
\begin{enumerate}
    \item \textbf{ROOAD dataset}. The   ROOAD dataset~\cite{chustz2021rooad} was collected using the Waterhog-UGV platform~\cite{clearpath_warthog}. The platform is equipped with a VectorNav VN-300 inertial navigation system (INS) \cite{vectornav_vn300} and an Ardusimple dual-antenna GNSS receiver with RTK capability for accurate position and heading estimation. The total duration of this dataset is approximately 45 minutes.
    \item \textbf{Hong Kong dataset}. the Hong Kong dataset~\cite{hsu2023hongkong} was collected using a passenger vehicle equipped with an Xsens MTi-10 \cite{xsens_migration} integrated with an RTK GNSS system. This dataset provides high-accuracy positioning and orientation measurements, with a total duration of approximately 100 minutes.
    \item \textbf{Our Dataset}. Our dataset was recorded by us using a Rosbot-XL \cite{husarion_rosbot_xl_manual} mobile robot platform. The robot is equipped with an Arazim Exiguo\textsuperscript{\textregistered} EX-300 \cite{arazim_ex300} and a dual-antenna GNSS system, which enables direct estimation of the vehicle heading. The total duration of this dataset is approximately 15 minutes.
\end{enumerate}
\noindent In total, the three datasets contain 160 minutes of real recorded inertial and GT measurements using three different mobile platforms with three different types of inertial sensors.
The complete specifications of the IMU sensors used in each dataset are provided in Table~\ref{imu_specification} and the carrying mobile platforms are shown in Fig. \ref{platform_datasets}. 
\begin{table}[h]
\centering
\footnotesize
\setlength{\tabcolsep}{4pt}
\caption{IMU sensor specifications for each of dataset}
\label{imu_specification}
\begin{tabular}{|c|cc|cc|}
\hline
                                                                         & \multicolumn{2}{c|}{\textbf{Gyroscope}}                                                                                                                                                 & \multicolumn{2}{c|}{\textbf{Accelerometer}}                                                                                                                                      \\ \hline
\textbf{\begin{tabular}[c]{@{}c@{}}Type \\ Inertial sensor\end{tabular}} & \multicolumn{1}{c|}{\textbf{\begin{tabular}[c]{@{}c@{}}Bias\\ {[}deg/h{]}\end{tabular}}} & \textbf{\begin{tabular}[c]{@{}c@{}}Noise\\ {[}deg/s/$\sqrt{\text{Hz}}${]}\end{tabular}} & \multicolumn{1}{c|}{\textbf{\begin{tabular}[c]{@{}c@{}}Bias\\ {[}mg{]}\end{tabular}}} & \textbf{\begin{tabular}[c]{@{}c@{}}Noise\\ {[}$\mu$g/$\sqrt{\text{Hz}}${]}\end{tabular}} \\ \hline
\textbf{VN-300 (ROOAD)}                                                  & \multicolumn{1}{c|}{5}                                                                   & 0.035                                                                                        & \multicolumn{1}{c|}{0.04}                                                             & 140                                                                                      \\ \hline
\textbf{Arazim EXL-300 (ours)}                                                 & \multicolumn{1}{c|}{10}                                                                  & 0.01                                                                                         & \multicolumn{1}{c|}{0.04}                                                             & 300                                                                                      \\ \hline
\textbf{MTi-10 (Honk-Kong)}                                              & \multicolumn{1}{c|}{18}                                                                  & 0.03                                                                                         & \multicolumn{1}{c|}{0.015}                                                            & 60                                                                                       \\ \hline
\end{tabular}
\end{table}
\subsection{Comparison with Existing Methods}
\noindent To evaluate the performance of the proposed approach, we compare it against three different methods:
\begin{itemize}
    \item \textbf{UKF}: The baseline UKF with GNSS updates, used to estimate the navigation state by fusing inertial measurements with GNSS position observations as described in Sections \ref{ins_gnss_fus} and \ref{ukf_sec}. In this approach the process and measurement noise covariances are constant throughout out the missions.
    \item \textbf{MB-AUKF}: A model-based adaptive UKF, in which the process noise covariance matrix ($\textbf{Q}$) and the measurement noise covariance matrix ($\textbf{R}$) are adaptively updated during filtering, as described in Section~\ref{model_base_adp}.
    \item \textbf{ANPN-UKF}: An adaptive neural process noise UKF, in which the process noise covariance matrix ($\textbf{Q}$) is estimated using $\sigma_{\textbf{Q}}$-Net as presented in Section~\ref{dnn_test}.
    \item \textbf{ANPMN-UKF (Ours)}: An adaptive neural UKF, where both the process noise covariance matrix ($\textbf{Q}$) and the measurement noise covariance matrix ($\textbf{R}$) are estimated using $\sigma_{\textbf{Q}}$-Net and $\sigma_{\textbf{R}}$-Net, respectively, as presented in Section~\ref{dnn_test}.
\end{itemize}
\subsection{Performance Metric}
\noindent To evaluate the proposed approach we used two performance metrics, as follows: 
\begin{enumerate}
    \item \textbf{Position root mean square error (PRMSE)}: The PRMSE of the 3D position compares the estimated position of the vehicle in the navigation frame with the GNSS-RTK GT:
    \begin{equation} \label{eq:rmse}
    \text{PRMSE (m)} = \sqrt{\frac{1}{N} \sum_{k=1}^{N} ||\textbf{p}_k-\hat{\textbf{p}}^{}_k||^{2}}
    \end{equation}
    where $\hat{\textbf{p}}_k$ are the 3D estimated position vector at time k and $\textbf{p}_k$ is the GT position vector at time k.
\end{enumerate}
\subsection{Results}
\noindent After training $\sigma_{\textbf{Q}}$-Net and $\sigma_{\textbf{R}}$-Net on the simulative data, we used the three datasets for evaluations using the PRMSE metric \eqref{eq:rmse}. Table \ref{Arazim-PRMSE} presents the PRMSE results for our dataset, which consists of five trajectories with varying lengths ranging from approximately 30 m to 40 m. The trajectories include different motion patterns and durations, providing a representative evaluation of filter robustness under realistic operating conditions. Table{~\ref{Arazim-PRMSE}} summarizes the PRMSE performance on the Arazim dataset. 
Overall, the proposed ANPMN-UKF achieves the lowest error across all five trajectories, reducing the average PRMSE to 2.56m compared to 2.90m (UKF), 2.82m (MB-AUKF), and 2.70m (ANPN-UKF).  This corresponds to an average improvement of 11.9{\%}, 9.2{\%}, and 5.3{\%} over UKF, MB-AUKF, and ANPN-UKF, respectively. 
%

\begin{table}[h]
\centering
\caption{Comparison of PRMSE across different trajectories in the Arazim dataset.}
\begin{tabular}{|ccccc|}

\hline
\multicolumn{5}{|c|}{\textbf{PRMSE {[}m{]}}}                                                                                                                                                                                                                                                 \\ \hline
\multicolumn{1}{|l|}{\textbf{No.Traj}}                         & \multicolumn{1}{l|}{\textbf{UKF}}                          & \multicolumn{1}{l|}{\textbf{MB-AUKF}}                         & \multicolumn{1}{l|}{\textbf{ANPN-UKF}}                        & \multicolumn{1}{l|}{\textbf{\begin{tabular}[c]{@{}c@{}}ANPMN-UKF\\ (ours)\end{tabular}}} \\ \hline
\multicolumn{1}{|c|}{\textbf{1}}                               & \multicolumn{1}{c|}{2.99}                                  & \multicolumn{1}{c|}{2.83}                                  & \multicolumn{1}{c|}{2.62}                                  & 2.48                                 \\ \hline
\multicolumn{1}{|c|}{\textbf{2}}                               & \multicolumn{1}{c|}{3.05}                                  & \multicolumn{1}{c|}{2.91}                                  & \multicolumn{1}{c|}{2.57}                                  & 2.46                                 \\ \hline
\multicolumn{1}{|c|}{\textbf{3}}                               & \multicolumn{1}{c|}{2.80}                                  & \multicolumn{1}{c|}{2.77}                                  & \multicolumn{1}{c|}{2.71}                                  & 2.51                                 \\ \hline
\multicolumn{1}{|c|}{\textbf{4}}                               & \multicolumn{1}{c|}{2.83}                                  & \multicolumn{1}{c|}{2.78}                                  & \multicolumn{1}{c|}{2.80}                                  & 2.66                                 \\ \hline
\multicolumn{1}{|c|}{\textbf{5}}                               & \multicolumn{1}{c|}{2.84}                                  & \multicolumn{1}{c|}{2.80}                                  & \multicolumn{1}{c|}{2.81}                                  & 2.67                                 \\ \hline

\multicolumn{1}{|l|}{\textbf{Average}} & \multicolumn{1}{c|}{\textbf{2.90}} & \multicolumn{1}{c|}{\textbf{2.82}} & \multicolumn{1}{c|}{\textbf{2.70}} & \textbf{2.56}                        \\ \hline
\end{tabular}
\label{Arazim-PRMSE}

\end{table}

\begin{figure}[h]
    \centering
    \includegraphics[width=1.0\linewidth]{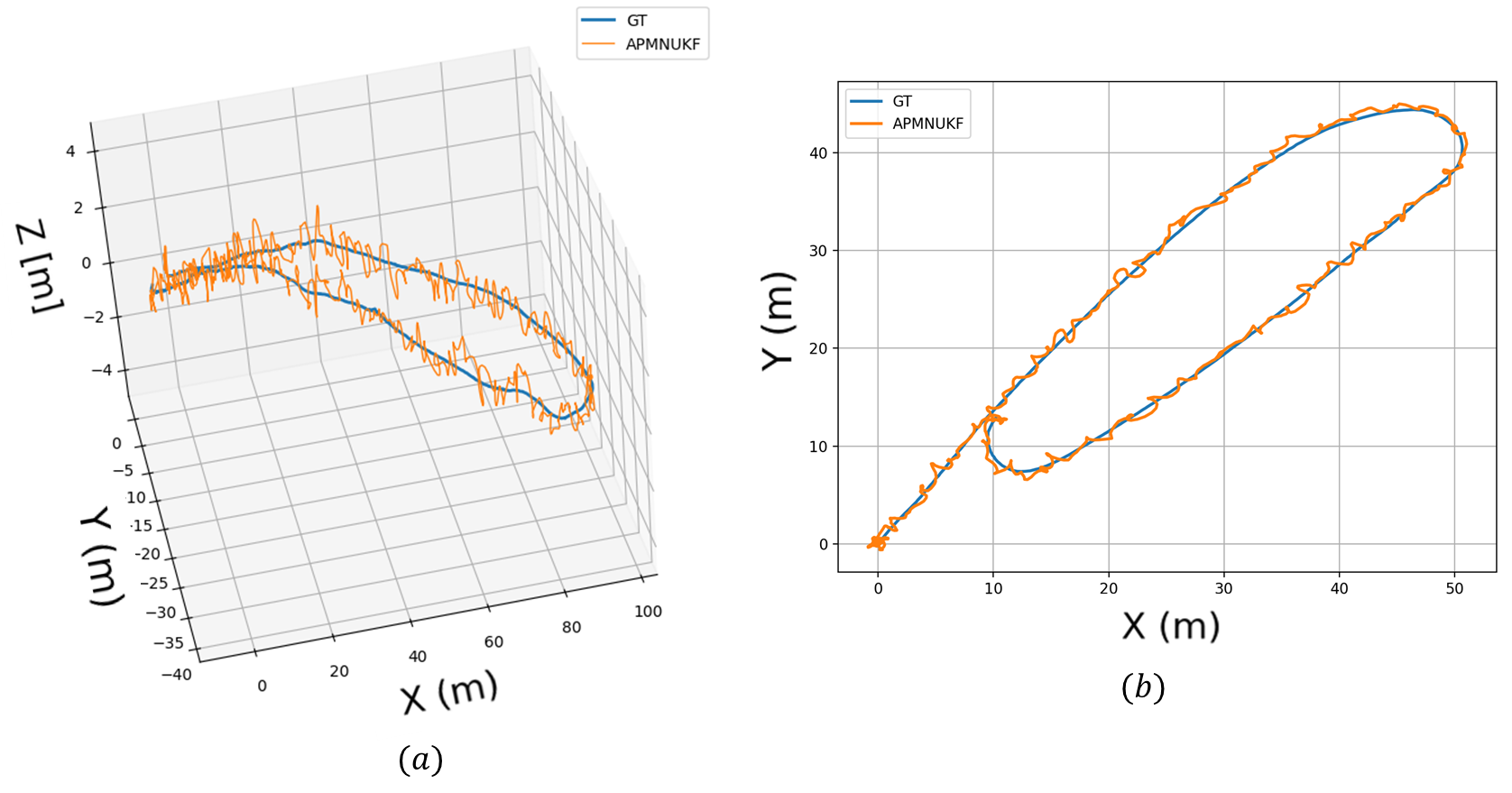}
    \caption{(a) 3D plot of Trajectory 1 showing the GT and our estimated trajectories. (b) 2D plot of Trajectory 1 showing the GT and our estimated trajectories.}       
    \label{trajctory_2d_3d}
\end{figure}

%
\noindent Table~\ref{ROOAD-PRMSE} presents the PRMSE results for the ROOAD dataset, which consists of six trajectories with varying lengths ranging from approximately 50 m to 130 m. The trajectories include different motions, up and down, and circle patterns. The ROOAD dataset was recorded in a sandy terrain environment, characterized by loose and deformable ground conditions. 
\noindent
Figure{~\ref{trajctory_2d_3d}} qualitatively shows  ANPMN-UKF (ours) and the GNSS ground-truth. Table{~\ref{ROOAD-PRMSE}} gives the results, showing our method achieves a PRMSE of 2.94m, improving  the UKF (4.69m), MB-AUKF (3.55m), and ANPN-UKF (3.45m).

\begin{table}[h]
\centering
\caption{Comparison of PRMSE across different trajectories in the ROOAD dataset.}

\begin{tabular}{|ccccc|}
\hline
\multicolumn{5}{|c|}{\textbf{PRMSE {[}m{]}}}                                                                                                                             \\ \hline
\multicolumn{1}{|c|}{\textbf{No.Traj}} & \multicolumn{1}{c|}{\textbf{UKF}}  & \multicolumn{1}{c|}{\textbf{MB-AUKF}} & \multicolumn{1}{c|}{\textbf{ANPN-UKF}} & \textbf{\begin{tabular}[c]{@{}c@{}}ANPMN-UKF\\ (ours)\end{tabular}} \\ \hline
\multicolumn{1}{|c|}{\textbf{1}}       & \multicolumn{1}{c|}{5.25}          & \multicolumn{1}{c|}{4.74}          & \multicolumn{1}{c|}{4.69}           & 3.04            \\ \hline
\multicolumn{1}{|c|}{\textbf{2}}       & \multicolumn{1}{c|}{5.58}          & \multicolumn{1}{c|}{3.73}          & \multicolumn{1}{c|}{3.68}           & 2.92            \\ \hline
\multicolumn{1}{|c|}{\textbf{3}}       & \multicolumn{1}{c|}{5.25}          & \multicolumn{1}{c|}{3.08}          & \multicolumn{1}{c|}{3.06}           & 2.97            \\ \hline
\multicolumn{1}{|c|}{\textbf{4}}       & \multicolumn{1}{c|}{5.28}          & \multicolumn{1}{c|}{3.77}          & \multicolumn{1}{c|}{3.33}           & 2.89            \\ \hline
\multicolumn{1}{|c|}{\textbf{5}}       & \multicolumn{1}{c|}{3.10}          & \multicolumn{1}{c|}{2.98}          & \multicolumn{1}{c|}{2.94}           & 2.94            \\ \hline
\multicolumn{1}{|c|}{\textbf{6}}       & \multicolumn{1}{c|}{3.66}          & \multicolumn{1}{c|}{3.02}          & \multicolumn{1}{c|}{2.99}           & 2.91            \\ \hline
\multicolumn{1}{|c|}{\textbf{Average}} & \multicolumn{1}{c|}{\textbf{4.69}} & \multicolumn{1}{c|}{\textbf{3.55}} & \multicolumn{1}{c|}{\textbf{3.45}}  & \textbf{2.94}   \\ \hline
\end{tabular}
\label{ROOAD-PRMSE}
\end{table}

\begin{figure}[h]
    \centering
    \includegraphics[width=1.0\linewidth]{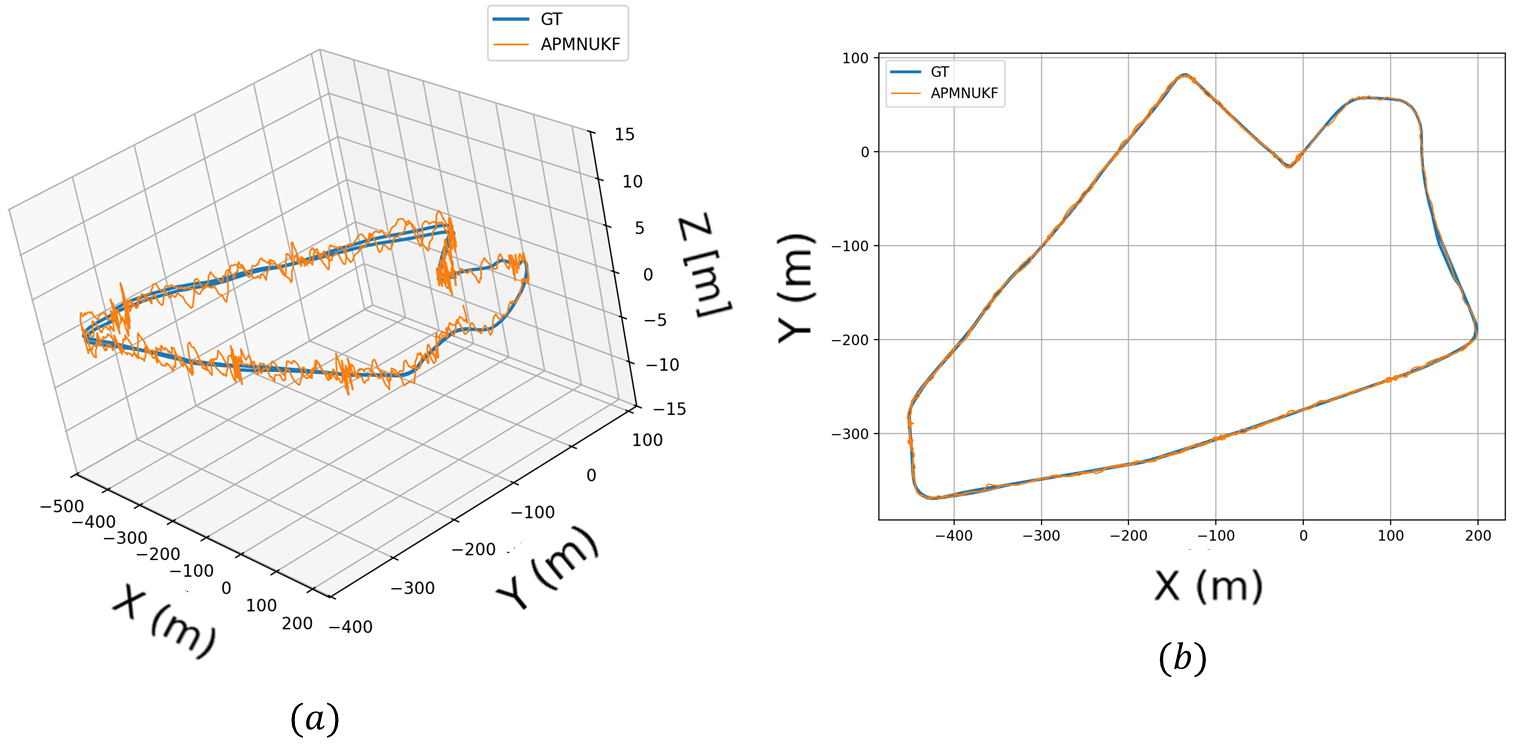}
    \caption{(a) 3D plot of Trajectory 2 showing the GT and our estimated trajectories. (b) 2D plot of Trajectory 2 showing the GT and our estimated trajectories.}
    \label{plot_1_hong_kong}
\end{figure}

\noindent Table \ref{Honk-Kong-PRMSE} presents the PRMSE results for the Hong-Kong dataset, which consists of four trajectories with varying lengths ranging from approximately 3Km m to 4 Km. The trajectories include different motions with car in the city of Hong-Kong. 
The GT and our estimated trajectories are shown in Fig.{\ref{plot_1_hong_kong}}. Table{~\ref{Honk-Kong-PRMSE}} shows that the ANPMN-UKF (ours) achieves a PRMSE of 3.07m, improving over UKF (3.79m), MB-AUKF (3.48m), and ANPN-UKF (3.21m). The gain is most pronounced in Trajectories 1 and 3, while in Trajectory 2 all methods perform similarly. In Trajectory 4 our approach remains close to the adaptive baselines and outperforms UKF.

\begin{table}[h]
\centering
\caption{Comparison of PRMSE across different trajectories in the Hong Kong dataset.}

\begin{tabular}{|ccccc|}
\hline
\multicolumn{5}{|c|}{\textbf{PRMSE {[}m{]}}}                                                                                                                             \\ \hline
\multicolumn{1}{|c|}{\textbf{No.Traj}} & \multicolumn{1}{c|}{\textbf{UKF}}  & \multicolumn{1}{c|}{\textbf{MB-AUKF}} & \multicolumn{1}{c|}{\textbf{ANPN-UKF}} & \textbf{\begin{tabular}[c]{@{}c@{}}ANPMN-UKF\\ (ours)\end{tabular}} \\ \hline
\multicolumn{1}{|c|}{\textbf{1}}       & \multicolumn{1}{c|}{5.32}          & \multicolumn{1}{c|}{4.45}          & \multicolumn{1}{c|}{3.31}           & 3.15            \\ \hline
\multicolumn{1}{|c|}{\textbf{2}}       & \multicolumn{1}{c|}{2.99}          & \multicolumn{1}{c|}{2.98}          & \multicolumn{1}{c|}{3.23}           & 2.98            \\ \hline
\multicolumn{1}{|c|}{\textbf{3}}       & \multicolumn{1}{c|}{3.62}          & \multicolumn{1}{c|}{3.45}          & \multicolumn{1}{c|}{3.17}           & 3.01            \\ \hline
\multicolumn{1}{|c|}{\textbf{4}}       & \multicolumn{1}{c|}{3.25}          & \multicolumn{1}{c|}{3.06}          & \multicolumn{1}{c|}{3.12}           & 3.16            \\ \hline
\multicolumn{1}{|c|}{\textbf{Average}} & \multicolumn{1}{c|}{\textbf{3.79}} & \multicolumn{1}{c|}{\textbf{3.48}} & \multicolumn{1}{c|}{\textbf{3.21}}  & \textbf{3.07}   \\ \hline
\end{tabular}
\label{Honk-Kong-PRMSE}
\end{table}

\subsection{Summary}
\noindent Using real-world data, we demonstrate that that our ANPMN-UKF achieves the highest position accuracy, consistently surpassing both the UKF, MB-AUKF, and ANPN-UKF. Specifically, the ANPMN-UKF improved the PRMSE estimation by 22.7\% relative to the UKF, by 12.72\% compared to the MB-AUKF, and by 8.03\% compared to the ANPN-UKF. Notably, both adaptive variants outperformed the baseline UKF in terms of position. The results are summarized in Table~\ref{tab:prmse_comparison} showing the average PRMSE across all trajectories and in Table~\ref{tab:improvement_comparison} the improvement using our approach over the baseline methods. 
\begin{table}[h]
\centering
\caption{Average PRMSE comparison across trajectories for the proposed and baseline methods.}
\begin{tabular}{|ccccc|}
\hline
\multicolumn{5}{|c|}{\textbf{Average PRMSE over all trajectories [m]}}           \\ \hline
\multicolumn{1}{|c|}{\textbf{Dataset}}   & \multicolumn{1}{c|}{\textbf{\begin{tabular}[c]{@{}c@{}}UKF \end{tabular}}} & \multicolumn{1}{c|}{\textbf{\begin{tabular}[c]{@{}c@{}}MB-AUKF \end{tabular}}} & \multicolumn{1}{c|}{\textbf{\begin{tabular}[c]{@{}c@{}}ANPN-UKF \end{tabular}}} & \textbf{\begin{tabular}[c]{@{}c@{}}ANPMN-UKF\\ (ours)\end{tabular}} \\ \hline
\multicolumn{1}{|c|}{\textbf{Hong-Kong}} & \multicolumn{1}{c|}{3.79}                                                            & \multicolumn{1}{c|}{3.48}                                                             & \multicolumn{1}{c|}{3.21}                                                              & 3.07                                                               \\ \hline
\multicolumn{1}{|c|}{\textbf{ROOAD}}     & \multicolumn{1}{c|}{4.69}                                                            & \multicolumn{1}{c|}{3.55}                                                             & \multicolumn{1}{c|}{3.45}                                                              & 2.94                                                               \\ \hline
\multicolumn{1}{|c|}{\textbf{Arazim}}    & \multicolumn{1}{c|}{2.90}                                                            & \multicolumn{1}{c|}{2.82}                                                             & \multicolumn{1}{c|}{2.70}                                                              & 2.56                                                               \\ \hline
\end{tabular}
\label{tab:prmse_comparison}

\end{table}
\begin{table}[h]
\centering
\caption{Improvement percentage comparison across different methods compare to our method ANPMN-UKF.}
\begin{tabular}{|cccc|}
\hline
\multicolumn{4}{|c|}{\textbf{ANPMN-UKF (ours) Improvement {[}\%{]}}}                                                                                                                           \\ \hline
\multicolumn{1}{|c|}{\textbf{Dataset}}   & \multicolumn{1}{c|}{\textbf{UKF}}   & \multicolumn{1}{c|}{\textbf{MB-AUKF}}  & \multicolumn{1}{c|}{\textbf{ANPN-UKF}}  \\ \hline
\multicolumn{1}{|c|}{\textbf{Hong-Kong}} & \multicolumn{1}{c|}{18.98}          & \multicolumn{1}{c|}{11.78}          & \multicolumn{1}{c|}{4.15}                        \\ \hline
\multicolumn{1}{|c|}{\textbf{ROOAD}}     & \multicolumn{1}{c|}{37.20}          & \multicolumn{1}{c|}{17.14}          & \multicolumn{1}{c|}{14.59}                       \\ \hline
\multicolumn{1}{|c|}{\textbf{Arazim}}    & \multicolumn{1}{c|}{11.92}          & \multicolumn{1}{c|}{9.25}           & \multicolumn{1}{c|}{5.35}                      \\ \hline
\multicolumn{1}{|c|}{\textbf{Average}}   & \multicolumn{1}{c|}{\textbf{22.70}} & \multicolumn{1}{c|}{\textbf{12.72}} & \multicolumn{1}{c|}{\textbf{8.03}}    \\ \hline
\end{tabular}
\label{tab:improvement_comparison}

\end{table}
\subsection{Computational Time}
\begin{figure}[h]
    \centering
    \includegraphics[width=1\linewidth]{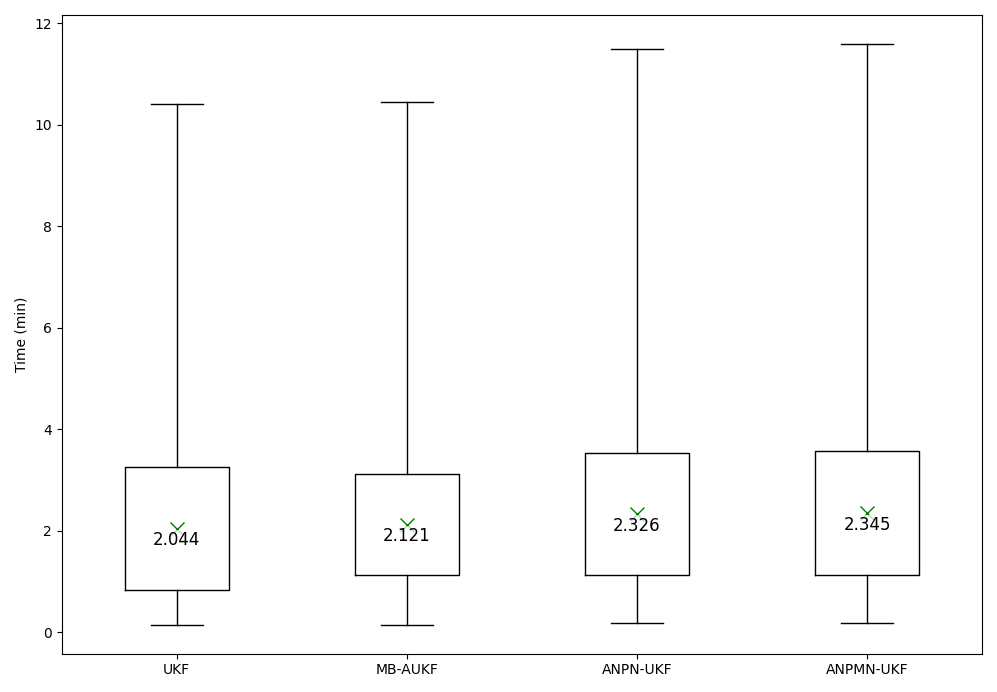}
    \caption{Run-time comparison across all datasets for UKF, MB-AUKF, ANPN-UKF, and ANPMN-UKF. The boxes indicate the average runtime, while the error bars represent the minimum and maximum values.}
    \label{runtime}
\end{figure}

\noindent Our code was implemented in Python using the PyTorch library. Training was performed on an NVIDIA GeForce RTX 5050 GPU 8GB. For evaluation, we used an Intel Core Ultra 7 255Hx CPU with 32\,GB RAM. We compare the run-time of our method against all baselines. In Fig.~\ref{runtime} presents the execution time for each approach across all trajectories and datasets; the boxes indicate the average runtime, while the error bars represent the minimum and maximum values.

\section{Conclusion}\label{sec:conclosuion} 
\noindent Reliable fusion of inertial and satellite-based navigation systems for ground mobile robots is highly dependent on accurate modeling of both process and measurement noise covariances. In practice, these noise characteristics are difficult to model precisely due to sensor imperfections, environmental effects, and changing motion dynamics. This work proposed a hybrid neural approach, ANPMN, that combines data-driven learning with classical estimation theory to improve noise uncertainty modeling within an UKF framework.\\
\noindent To this end, we proposed a neural architecture backbone for estimating the inertial uncertainty, $\sigma_{\textbf{Q}}$-Net, and for estimating the position uncertainty, $\sigma_{\textbf{R}}$-Net. The neural network is introduced to estimate the process noise covariance and measurement noise covariance parameters directly from raw inertial sensor and GNSS measurements. The network operates without requiring external features or handcrafted tuning parameters, enabling end-to-end noise characterization based solely on sensor data. \\
%
\noindent Using a sim2real approach,  we evaluated our ANPMN-UKF methods, on three independent datasets collected using different inertial sensors and platforms which operated in different conditions. Our sim2real approach enables training on large scale data, while reducing the need for costly and time-consuming real-world data collection. Experimental results demonstrate consistent performance improvements across all tested trajectories. On average, our proposed approach improves positioning accuracy in terms of the PRMSE by 22.7\% compared to a standard UKF, by 12.72\% compared to an MB-AUKF, and by 8.03\% compared to an ANPN-UKF, across all datasets. \\
\noindent These results indicate that the learned noise model generalizes well to unseen datasets, inertial sensors, road surfaces, and environmental conditions, demonstrating robustness beyond the simulative training distribution. Although experimental validation focused on wheeled platforms, the proposed framework is not platform-specific and can be readily applied to any inertial navigation system augmented with external measurements.
A main limitation of our approach is that the learned noise model may degrade when real-world conditions deviate significantly from the simulated training setup. In addition, using the deep learning for estimate the noise adaptation increases the per-iteration computational cost compared to a standard UKF. 
\noindent Overall, this study establishes a scalable methodology (sim2real and ANPMN-UKF) for embedding deep learning into adaptive state estimation using the UKF, laying a solid foundation for future research in robust and accurate navigation systems across a wide range of operating scenarios.
\section*{Acknowledgment}
\noindent
The authors gratefully acknowledge the support of the Israel Innovation Authority under grant 84330, which partially supported this research.\\
The authors would like to thank Arazim Ltd. for their support and for providing the Arazim Exiguo EX-300 unit used in the data collection.

\bibliographystyle{ieeetr}
\bibliography{bio}

\end{document}